\documentclass[notoc]{kling-keye}
\usepackage{amsfonts}
\usepackage{newtxmath} 
\usepackage{multirow}
\usepackage{colortbl}
\usepackage{pifont}
\usepackage{xspace}
\usepackage{placeins}
\usepackage{needspace}
\definecolor{aceteal}{HTML}{246D73}
\definecolor{aceochre}{HTML}{B77C35}

\newcommand{\bench}{VideoACE\xspace}

\newcommand{\relook}{video harness\xspace}
\newcommand{\single}{\textsc{Single}\xspace}

\usepackage{pgfplots}
\usetikzlibrary{patterns}
\pgfplotsset{compat=1.18}
\definecolor{aceblue}{HTML}{2166AC}
\definecolor{acered}{HTML}{B2182B}

\patchcmd{\maketitle}{\fontsize{19}{23}}{\fontsize{17}{21}}{}{\PackageError{videoace}{Title size patch failed}{Check the title template.}}

\title{VideoACE: Benchmarking Agentic Video Understanding through High-Frame-Rate Actions}
\author{{\normalfont\begin{tabular}{@{}c@{\hspace{2em}}c@{\hspace{2em}}c@{}}Luoyuan Zhang\textsuperscript{*,\textsection} & Xuanyu Zheng\textsuperscript{*,\textdagger,\textdaggerdbl} & Yancheng Long\textsuperscript{*} \\[3pt] Haoyang Xu\textsuperscript{*} & Bin Wen\textsuperscript{\textdagger} & Tingting Gao\end{tabular}}}
\affiliation{KlingAI Research}
\headertext{VideoACE: Agentic Video Understanding}
\reportdate{November 18, 2026}
\projectlinks{}
\reportabstract{Agentic video understanding requires models to actively interact with video environments to gather evidence and form judgments. Evaluating this capability therefore requires assessing whether models can translate opportunities for interaction into better understanding. We introduce \bench, a benchmark for assessing video understanding under conditions that support sustained interaction. We focus on high-frame-rate action videos, where rapid, subtle changes make targeted observation especially valuable. The benchmark comprises 1,000 multiple-choice questions spanning six categories of fine-grained action understanding, accompanied by temporal evidence annotations and user hints screened for answer leakage. Its evaluation environment combines video inspection tools with a user simulator, allowing models to adjust temporal coverage, sampling density, and spatial detail, and to revisit evidence and revise their judgments under simulated user feedback. We evaluate single-pass observation, autonomous tool use, and tool use combined with user feedback to examine how effectively models exploit these opportunities. Evaluation of five models reveals distinct patterns in how they acquire evidence and respond to opportunities for further interaction. These findings establish VideoACE as a testbed for examining agentic video understanding beyond performance under a fixed observation setting.
}
\begin{document}
\maketitle
\begingroup
\renewcommand{\thefootnote}{\fnsymbol{footnote}}
\footnotetext[1]{Equal contribution.}
\footnotetext[2]{Corresponding authors: Xuanyu Zheng and Bin Wen.}
\footnotetext[3]{Project lead: Xuanyu Zheng.}
\footnotetext[4]{This work was conducted during the author's internship at KlingAI.}
\endgroup
\begin{figure}[ht]
\centering
\includegraphics[width=\linewidth]{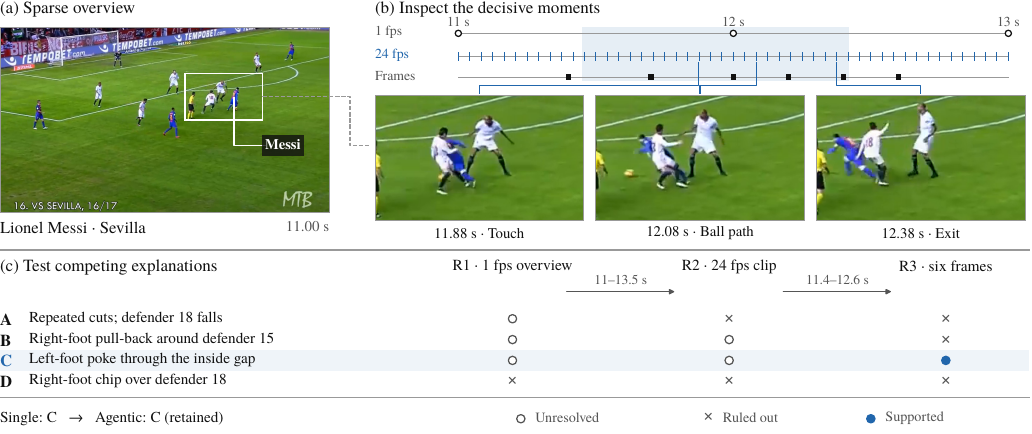}
\captionsetup{hypcap=false}
\caption{Targeted video inspection for understanding a fast action.}
\captionsetup{hypcap=true}
\label{fig:teaser}
\end{figure}

\section{Introduction}
\label{sec:intro}

Agentic video understanding requires models to actively interact with video environments to gather evidence and form judgments. Models must decide when and where to look, at what temporal density and spatial scale, and how to use the resulting evidence to understand the video. Evaluating this capability requires an interactive video environment and tasks that test how well models use interaction.

To give models sufficient opportunity to demonstrate this capability, the evaluation environment should support further evidence gathering, responses to feedback, and refinement of judgments beyond the initial observation. Video tools enable active inspection, while user feedback can encourage models to examine aspects of the video that remain unresolved. Incorporating both into the benchmark allows models to work through a sequence of observations and feedback, enabling us to assess whether these opportunities translate into better understanding.

We introduce \bench\ to study this question. We focus on fast, fine-grained actions, where decisive evidence can be brief and visually subtle. For example, a sparse view of a football dribble may reveal a player's position before and after a challenge while leaving the intervening touch unresolved. Answering a question about that touch may require a denser view of the relevant interval or a closer inspection of the contact region. Such scenes make the choice of observation timing, sampling density, and spatial coverage consequential, providing concrete tasks for evaluating agentic video understanding.

We curate 100 videos from a large pool of high-resolution, high-frame-rate videos available online and construct 1,000 multiple-choice questions spanning six categories of fine-grained action understanding. Videos are retained at 24\,fps, and questions are accompanied by temporal evidence annotations with a median window duration of 0.90\,s. The construction process includes question generation, option refinement, automated screening, and human review and validation, grounding questions in specific action details and their visual evidence. Evidence annotations also support the construction of user hints, which are screened for answer leakage to provide task-relevant guidance during interaction.

To support further exploration of this evidence, \bench\ provides video inspection tools that let models adjust temporal coverage, sampling density, and spatial detail. Models can revisit clips, select frames, or inspect local regions as needed. \Cref{fig:teaser} illustrates this process with a separately constructed example outside the test set. Building on this tool access, a User Simulator supplies feedback during interaction: it uses the evidence regions reported by the model to provide temporal pointers and visual focus cues, encouraging further inspection without revealing whether an answer is correct. Together, the tools and the simulator form an evaluation environment that supports sustained interaction. VideoACE examines how models use these tools, respond to feedback, and complete the tasks within this environment.

Experiments with five models show substantial differences in the gains they obtain from opportunities for interaction. Models differ in whether they pursue further observations and how effectively they use the resulting evidence. These differences highlight the importance of evaluating how effectively models use the observations made available by tools. Through its fine-grained action questions, interactive video environment, and analyses of model performance, \bench\ provides a benchmark for systematically assessing this capability.

\section{Related Work}
\label{sec:related}

\paragraph{Evaluating video understanding.}
Early video question answering benchmarks such as TVQA~\cite{tvqa}, ActivityNet-QA~\cite{activitynetqa}, and NExT-QA~\cite{nextqa} test event recognition and temporal and causal reasoning. Video-MME~\cite{videomme}, MVBench~\cite{mvbench}, and Video-MME-v2~\cite{videommev2} evaluate a broad range of capabilities for multimodal models~\cite{gemini15,qwen25vl,videollava,qwen3vl,qwen3omni,mimovl,step3,yu2026minicpm,cui2026minicpm,zhang2026liberating,kwaikeyeteam2026kwaikeyevl20technicalreport}. EgoSchema~\cite{egoschema}, MLVU~\cite{mlvu}, and LongVideoBench~\cite{longvideobench} extend evaluation to long videos, while Perception Test~\cite{perceptiontest} and TempCompass~\cite{tempcompass} probe perceptual and temporal skills. All of them score answers from a fixed observation of the video. \bench instead lets models request further observations and respond to simulated user feedback, and measures whether these opportunities raise accuracy.

\paragraph{Fine-grained motion.}
Something-Something~\cite{ssv2}, Diving48~\cite{diving48}, and FineGym~\cite{finegym} show that recognizing fine-grained actions requires temporal detail beyond scene appearance, and SlowFast~\cite{slowfast} captures fast motion with a high-frame-rate pathway. MotionBench~\cite{motionbench} and FAVOR-Bench~\cite{favorbench} bring these concerns to multimodal language models and report benefits from higher-rate inputs. \bench uses fast, fine-grained actions so that observation timing, density, and spatial detail affect the answer.

\paragraph{Adaptive visual observation.}
ReAct~\cite{react} interleaves reasoning with actions in an environment, and VisProg~\cite{visprog} and ViperGPT~\cite{vipergpt} compose vision tools to answer visual queries. V*~\cite{vstar} searches high-resolution images for small targets. For video, VideoAgent~\cite{videoagent} iteratively gathers information from long videos, VideoTree~\cite{videotree} builds a query-adaptive hierarchical representation, and EVA~\cite{eva} trains an end-to-end video agent with reinforcement learning. VideoNexus~\cite{long2026videonexusvideoperceptionreasoning} separates video perception from planning and reasoning, using perception sub-agents to return compact, temporally grounded evidence. \bench evaluates how models use such observation opportunities in a shared environment for fast actions.

\paragraph{Interactive agent evaluation.}
AgentBench~\cite{agentbench} and WebArena~\cite{webarena} evaluate language agents inside interactive environments. MINT~\cite{mint} adds language feedback across multiple turns of tool use, and $\tau$-bench~\cite{taubench} simulates users who interact with tool-using agents. \bench brings this setting to video, combining inspection tools with a User Simulator that points models to relevant evidence.

\section{Dataset Construction}
\label{sec:construction}

The dataset consists of questions whose answers depend on brief visual evidence. Construction proceeds in four stages: selecting observable action scenes, generating questions about short action details, validating the question set, and preparing feedback for further inspection.

\subsection{Selecting videos for detailed observation}
\label{sec:videos}

We curate 100 videos from a large pool of high-resolution, high-frame-rate videos available online. The source library contains 800 videos spanning film fights, combat sports, martial-arts demonstrations, team sports, fast non-contact sports, dance, drama action, single-person motion, and animation or game footage. Retrieval uses category-specific queries, deduplicated source identifiers, and title filters. Successfully probed sources of at least 720p enter window selection.

Window selection scores visible-person occupancy and image motion, and a cap of six clips per channel limits source concentration. Appendix~\ref{sec:supp-pipeline} lists the exact thresholds, rejection counts, and provenance records.

The evaluation clips are supplied at 24\,fps with audio removed. Durations range from 35.5 to 59.9\,s (mean 57.4\,s).

\paragraph{Making actions observable.}
A window of at most 59.5\,s is scored as $0.6p+0.4\min(m/15,1)$, where $p$ is the fraction of sampled frames meeting the category's person requirement and $m$ is mean grayscale frame difference. Multi-person scenes require at least two visible people in half the samples; single-person scenes require one in 70\%. A motion threshold of 1.5 removes nearly static candidates. The selected clips keep participants in view.

Category-proportional sampling preserves the library's composition across action domains. The 100-video manifest uses largest-remainder allocation and seeded sampling within each category.

\subsection{Questions grounded in short action details}
\label{sec:generation}\label{sec:types}\label{sec:annotations}

Each video contributes ten four-choice questions covering actor--action binding, interactions, temporal order, counting, spatial relations, and transient motion details. The options share the coarse before and after states and differ in an intermediate action, such as the contacting limb, recipient, direction, count, or order.

Gemini 4 Argon first captions the silent video at 24\,fps with high reasoning effort and high media resolution, then generates questions in the same conversation. An earlier Gemini 3.6 Flash caption produced at 8\,fps supplies candidate mistakes to check against the denser observations. Distractors include swapped actors, reversed directions, altered counts or order, and mistakes suggested by that caption. Each item records its answer, evidence interval, rationale, distractor explanations, and an estimated temporal-density label in $\{8,24\}$\,fps, which we use for difficulty analysis.

\begin{table}[t]
\centering
\caption{\textbf{Six action reasoning skills.} Counts refer to questions; each question is assigned one primary type.}
\label{tab:types}
\small
\setlength{\tabcolsep}{3pt}
\renewcommand{\arraystretch}{1.12}
\begin{tabular}{@{}lp{3.4cm}r@{}}
\toprule
Type & What must be resolved & $n$ \\
\midrule
Binding & Which person acts & 75 \\
Interaction & Who acts on whom & 165 \\
Order & Which action comes first & 227 \\
Counting & How often an action repeats & 84 \\
Spatial & Where and in which direction & 132 \\
Transient & Which brief motion occurs & 317 \\
\bottomrule
\end{tabular}
\end{table}

\paragraph{Connecting task types to observation needs.}
The six categories in \cref{tab:types} place different demands on observation. Binding and interaction require tracking participants and their roles; order and counting require tracking a sequence of actions; spatial relations and transient details require resolving direction or a short-lived configuration. Each question carries one primary type.

The distractors keep the surrounding event and change its decisive detail. For example, question y001-4 distinguishes a right-hand rod moving upward followed by a left-hand rod moving downward from the reversed order of those same actions. Its annotated evidence spans 10.45--10.93\,s. Recognizing the participants and the overall event leaves this distinction open; the model must resolve the temporal relation between the two actions.

\subsection{Quality control and human validation}
\label{sec:leakage}

Quality control combines automated option refinement with human review and validation. A DeepSeek-V4.1-Flash stage rewrites all four options while retaining the key and each distractor's intended error, with constraints on relative length, sentence structure, and descriptive detail. Of 1,000 questions, 991 complete rewriting in one round and nine require a second. Answer positions are then balanced to 250 instances of each letter.

After rewriting, the fraction of questions whose correct option is strictly the longest falls from 42.7\% to 8.5\%, and the mean relative length spread falls from 0.205 to 0.038. Appendix~\ref{sec:supp-pipeline} gives the full comparison.

\paragraph{Separating visual distinctions from option form.}
If the correct choice systematically contains more words or qualifications, a model can answer from the wording without inspecting the action. The refinement stage therefore keeps each option's visual claim and equalizes its surface form. After rewriting, only 317 questions have a unique longest option, so a uniformly placed key would be strictly longest in 7.93\% of questions, close to the observed 8.5\%.

\subsection{Preparing feedback for further interaction}
\label{sec:feedback-data}

The evidence annotations also provide the material for user feedback. A preprocessing stage removes temporal references from the question stems, keeps the options and keys, and prepares coarse and fine position hints and visual focus cues. Position hints point to a time region; visual cues name a participant, body part, or object to examine.

Hints are released into a conversation shared by the question group, so screening covers the target question and the other questions in its group. Text-only checks reject cues that reveal answer-specific information. The task artifact provides position hints for 999 questions and visual focus cues for 682. Appendix~\ref{sec:supp-guided} details the screening policy.

\subsection{Dataset composition}
\label{sec:stats}

The benchmark contains 1,000 questions over 5,744\,s of footage. Evidence windows average 0.93\,s and have a median of 0.90\,s. Because most evidence is this short, the environment lets models refine observations in time and space.

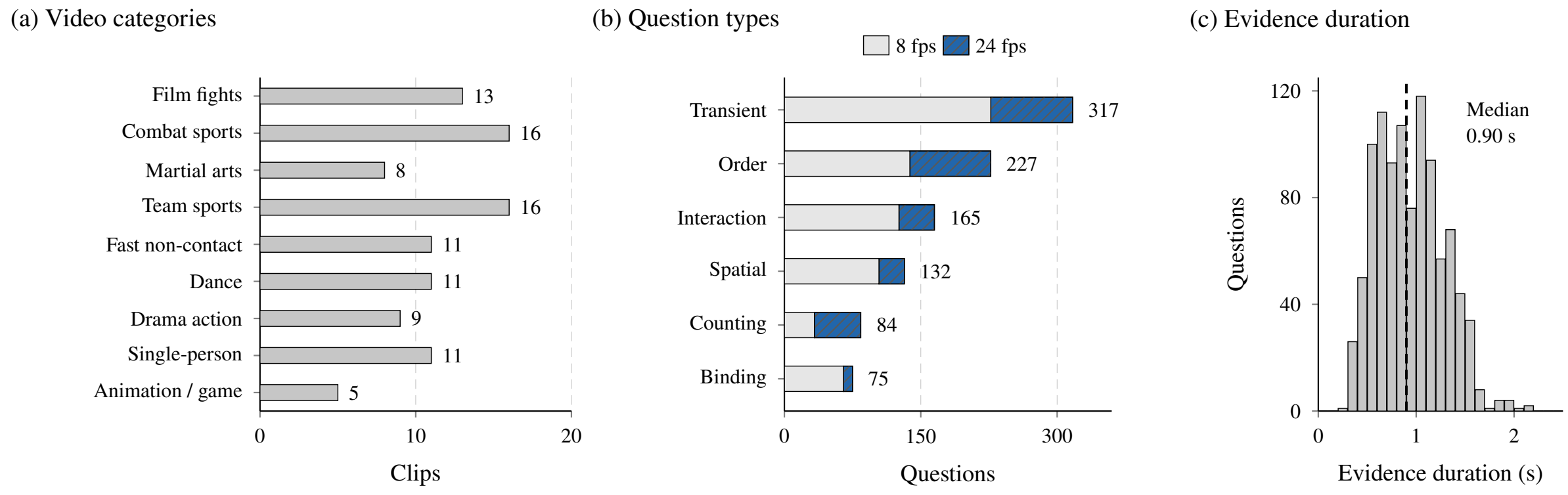
\begin{figure*}[t]
\centering
\resizebox{\linewidth}{!}{\pgfplotsset{aceaxis/.style={scale only axis,axis lines*=left,axis line style={line width=.45pt},tick style={line width=.45pt},tick align=outside,major tick length=2pt,font=\fontsize{7}{8}\selectfont,label style={font=\fontsize{7.5}{9}\selectfont},tick label style={font=\fontsize{6.5}{8}\selectfont},grid style={gray!40,densely dashed,line width=.25pt},clip=false},acebar/.style={draw=black,line width=.3pt}}
\begin{tikzpicture}[x=1pt,y=1pt]
\path[use as bounding box] (0,0) rectangle (495,156);
\begin{axis}[aceaxis,at={(82pt,26pt)},anchor=south west,width=98pt,height=105pt,xbar,bar width=5pt,xmin=0,xmax=20,ymin=-.5,ymax=8.5,y dir=reverse,ytick={0,...,8},yticklabels={Film fights,Combat sports,Martial arts,Team sports,Fast non-contact,Dance,Drama action,Single-person,Animation / game},xtick={0,10,20},xlabel={Clips},xmajorgrids]
\addplot[acebar,fill=gray!45,nodes near coords,nodes near coords style={font=\fontsize{6.5}{7}\selectfont},point meta=x] coordinates {(13,0) (16,1) (8,2) (16,3) (11,4) (11,5) (9,6) (11,7) (5,8)};
\end{axis}
\begin{axis}[aceaxis,at={(247pt,26pt)},anchor=south west,width=103pt,height=105pt,xbar stacked,bar width=8pt,xmin=0,xmax=360,ymin=-.6,ymax=5.6,y dir=reverse,ytick={0,...,5},yticklabels={Transient,Order,Interaction,Spatial,Counting,Binding},xtick={0,150,300},xlabel={Questions},xmajorgrids,legend style={draw=none,at={(.5,1.02)},anchor=south,legend columns=2,font=\fontsize{6.5}{8}\selectfont},legend cell align=left]
\addplot[acebar,fill=gray!20] coordinates {(227,0) (138,1) (126,2) (104,3) (33,4) (65,5)};
\addplot[acebar,fill=aceblue,postaction={pattern=north east lines,pattern color=black!65}] coordinates {(90,0) (89,1) (39,2) (28,3) (51,4) (10,5)};
\legend{8 fps,24 fps}
\node[anchor=west,font=\fontsize{6.5}{8}\selectfont] at (axis cs:322,0) {317};
\node[anchor=west,font=\fontsize{6.5}{8}\selectfont] at (axis cs:232,1) {227};
\node[anchor=west,font=\fontsize{6.5}{8}\selectfont] at (axis cs:170,2) {165};
\node[anchor=west,font=\fontsize{6.5}{8}\selectfont] at (axis cs:137,3) {132};
\node[anchor=west,font=\fontsize{6.5}{8}\selectfont] at (axis cs:89,4) {84};
\node[anchor=west,font=\fontsize{6.5}{8}\selectfont] at (axis cs:80,5) {75};
\end{axis}
\begin{axis}[aceaxis,at={(415pt,26pt)},anchor=south west,width=77pt,height=105pt,xmin=0,xmax=2.5,ymin=0,ymax=125,xtick={0,1,2},ytick={0,40,80,120},xlabel={Evidence duration (s)},ylabel={Questions}]
\addplot[ybar,bar width=2.9pt,acebar,fill=gray!45] coordinates {(0.05,0) (0.15,0) (0.25,1) (0.35,26) (0.45,50) (0.55,100) (0.65,112) (0.75,93) (0.85,107) (0.95,76) (1.05,118) (1.15,94) (1.25,57) (1.35,68) (1.45,44) (1.55,34) (1.65,8) (1.75,1) (1.85,4) (1.95,4) (2.05,1) (2.15,2) (2.25,0) (2.35,0) (2.45,0)};
\addplot[black,densely dashed,line width=.8pt] coordinates {(0.9,0) (0.9,125)};
\node[anchor=north west,align=left,font=\fontsize{6.5}{8}\selectfont] at (axis cs:1.4,120) {Median\\0.90 s};
\end{axis}
\node[anchor=west,font=\fontsize{8}{9}\selectfont] at (0,149) {(a) Video categories};
\node[anchor=west,font=\fontsize{8}{9}\selectfont] at (183,149) {(b) Question types};
\node[anchor=west,font=\fontsize{8}{9}\selectfont] at (371,149) {(c) Evidence duration};
\end{tikzpicture}}
\caption{The action tasks used to evaluate interaction. Left: source clip categories. Center: question counts by estimated temporal density. Right: annotated evidence-window lengths, with a median of 0.90\,s.}
\label{fig:stats}
\end{figure*}

\section{Interactive Evaluation Environment}
\label{sec:design}

The environment offers models two forms of interaction: tools for acquiring visual evidence and simulated user feedback that prompts further inspection. The evaluation measures how models use both to answer the action questions.

\begin{figure*}[t]
\centering
\includegraphics[width=\textwidth]{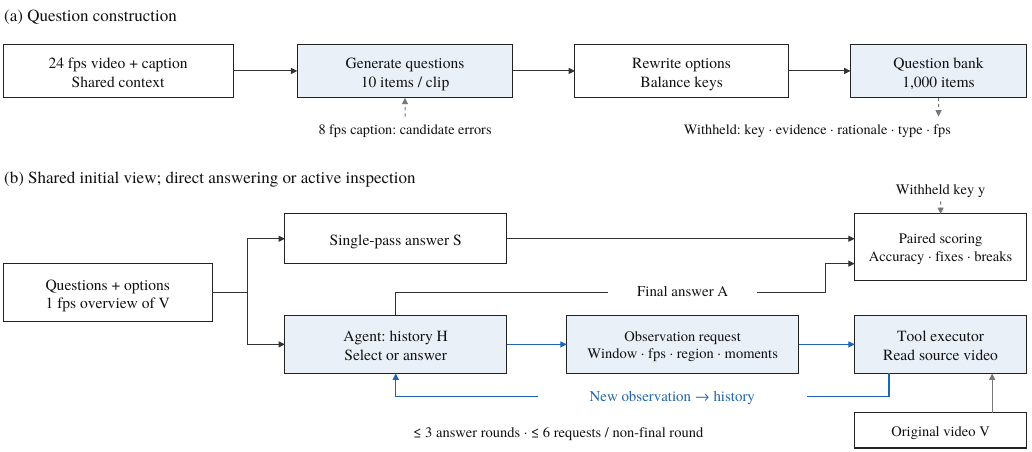}
\caption{Question construction and the autonomous observation path. (a) Dense video and captions support question generation; a weaker caption supplies candidate mistakes. Option refinement controls surface form and balances answer positions. (b) Starting from a sparse overview, a model requests further observations and accumulates evidence before answering. The User Simulator (\cref{sec:guided-protocol}) extends this interaction through feedback.}
\label{fig:pipeline}
\end{figure*}

\subsection{Video, model, and interaction}
\label{sec:measurement}

Given a source video $V$ and questions $Q$, the model receives an initial observation $O_0=\mathcal{S}_{1\mathrm{fps}}(V)$. The video is the environment, the model is the agent, and the observation tools mediate between them. Requests depend on the questions and the interaction history:
\begin{equation}
 a_t=\pi(Q,H_t),\qquad O_{t+1}=\mathcal{T}(V,a_t),
\end{equation}
where $H_t$ contains the available observations, model responses, and any user feedback, and $\mathcal{T}$ is the shared tool executor. The model decides which evidence to seek and how to use it in its answer.

For fast actions, the 1\,fps overview shows the scene but leaves a brief contact or change of direction unresolved. Additional observations let the model inspect that evidence at a suitable temporal density or spatial scale.

\subsection{Tools for acquiring visual evidence}
\label{sec:protocols}

The executor supports six operations: temporal clips at up to 24\,fps, selected still frames, spatial crops, numbered occurrence montages, broad 2\,fps scans, and a refreshed overview. With these operations, a model controls when and where it looks in the original video.

Each inner tool session begins with a 1\,fps overview and permits at most three answer rounds. In each non-final round, the model may request up to six observations. A selected-frame request returns at most eight images; a dense clip is adjusted to a soft budget of 120 frames. Observations are appended to the inner conversation, so later requests build on the evidence already obtained. The evaluated model selects observations and answers by itself, with no separate planning model.

\subsection{User feedback supporting further inspection}
\label{sec:guided-protocol}

The User Simulator stands in for a user who points the model to relevant evidence. It supplies temporal pointers and visual focus cues that prompt the model to inspect the video again. Its online policy is deterministic: it selects from the prepared hints according to the evidence windows the model reports, with no additional language-model call.

At each outer turn, the model answers the group of questions and reports up to three supporting intervals per question. If those intervals miss the annotated evidence, the simulator provides a coarse position hint (H1), followed by a tighter hint (H2). Once an interval meets the overlap rule, or the position hints are exhausted, a visual focus cue (H3) names a participant, body part, or object. The model then uses its tools again and may revise its answer. Interaction ends when no new hint remains or after four outer turns.

Hint release never depends on answer correctness, and the simulator never reveals whether a selected option is correct. Appendix~\ref{sec:supp-guided} specifies hint release and screening.

\paragraph{From reported evidence to the next interaction.}
For annotated interval $E_i$ of length $\ell_i$, a reported window $W$ counts as located when
\begin{equation}
\frac{|W\cap E_i|}{\max(\ell_i,10^{-3})}\geq 0.5,
\quad |W|\leq\max(10\,\mathrm{s},3\ell_i).
\label{eq:location}
\end{equation}
The coverage condition accepts approximate localization, and the length bound keeps a whole-video report from satisfying it. H1 provides a coarse window of nominal length $\max(12\,\mathrm{s},2\ell_i)$, and H2 rounds the evidence boundaries outward to 0.5\,s. H3 shifts the focus from time to a visual entity. The episode stops when no hints remain.

\subsection{Evaluation settings and reporting}
\label{sec:metrics}

Questions are grouped four, three, and three per video, yielding 300 groups. We evaluate three settings.

\paragraph{Single-pass observation.}
\single receives the complete clip at 1\,fps and answers once.

\paragraph{Autonomous tool use.}
The model receives the same overview through the video harness and uses the inner tool session to inspect and answer. This setting and the single-pass setting both use the original question stems, including their time references. Evidence annotations and answer keys are withheld.

\paragraph{Tool use with user feedback.}
The model uses the same tool backend within each outer turn and receives simulator feedback between turns. This setting uses the prepared stems from \cref{sec:feedback-data}. Each outer turn starts a new inner tool session with the cumulative per-question hints; visual history is kept within a session and reset between outer turns.

For autonomous tool use, we report full-set accuracy and the gain over single-pass observation. For the feedback setting, we report first- and final-outer-turn accuracy on the same question set.

\section{Experiments}
\label{sec:experiments}

\subsection{Evaluation setup}
\label{sec:setup}

\begin{table*}[t]
\centering
\caption{\textbf{Autonomous tool use on \bench.} Each model answers all 1,000 questions. Gain is the accuracy difference in percentage points. The shared interface provides the same operations, but models differ in the benefit they obtain.}
\label{tab:main}
\small
\setlength{\tabcolsep}{16pt}
\renewcommand{\arraystretch}{1.18}
\begin{tabular}{@{}lrrr@{}}
\toprule
Model & Single-pass (\%) & Tool use (\%) & Gain \\
\midrule
Gemini 4 Argon & \textbf{80.8} & \textbf{88.1} & +7.3 \\
Gemini 3.8 Flash & 75.4 & 86.4 & +11.0 \\
Gemini 3.6 Flash & 65.6 & 79.8 & +14.2 \\
\addlinespace[3pt]
Qwen3.8-Flash & 63.5 & 76.2 & +12.7 \\
Qwen3.7-Flash & 56.3 & 59.3 & +3.0 \\
\bottomrule
\end{tabular}
\end{table*}

\begin{table*}[t]
\centering
\caption{\textbf{Round usage and recorded cost.} R1 is first-round agentic accuracy (\%); deferred answers count as incorrect. Round counts sum to 300 groups. Costs (USD) are logged estimates for completed records.}
\label{tab:cost}
\small
\setlength{\tabcolsep}{7pt}
\renewcommand{\arraystretch}{1.15}
\begin{tabular}{@{}lrrr@{\hspace{18pt}}rr@{}}
\toprule
Model & R1 & Deferred & Rounds 1 / 2 / 3 & Single (\$) & Agentic (\$) \\
\midrule
Gemini 4 Argon & 79.9 & 0 & 1 / 266 / 33 & 12.81 & 72.39 \\
Gemini 3.8 Flash & 71.8 & 18 & 1 / 152 / 147 & 5.76 & 26.96 \\
Gemini 3.6 Flash & 64.1 & 0 & 1 / 279 / 20 & 3.02 & 16.19 \\
Qwen3.8-Flash & 60.1 & 23 & 0 / 101 / 199 & 2.96 & 11.59 \\
Qwen3.7-Flash & 53.8 & 59 & 167 / 124 / 9 & 1.68 & 3.48 \\
\bottomrule
\end{tabular}
\end{table*}

We evaluate Gemini 4 Argon, Gemini 3.8 Flash, Gemini 3.6 Flash, Qwen3.8-Flash, and Qwen3.7-Flash via API. All models use the same source videos, question bank, tool executor, and request limits. Gemini uses high reasoning effort; Qwen uses the gateway's default reasoning behavior, temperature 0.1, and a maximum of 8,192 output tokens. Gemini 4 Argon also generates the questions (\cref{sec:generation}).

After retries, every model completes all 300 groups with 1,000 valid predictions in all three settings (\cref{tab:main,tab:combined-feedback}).

\subsection{Do models benefit from autonomous observation?}
\label{sec:main}

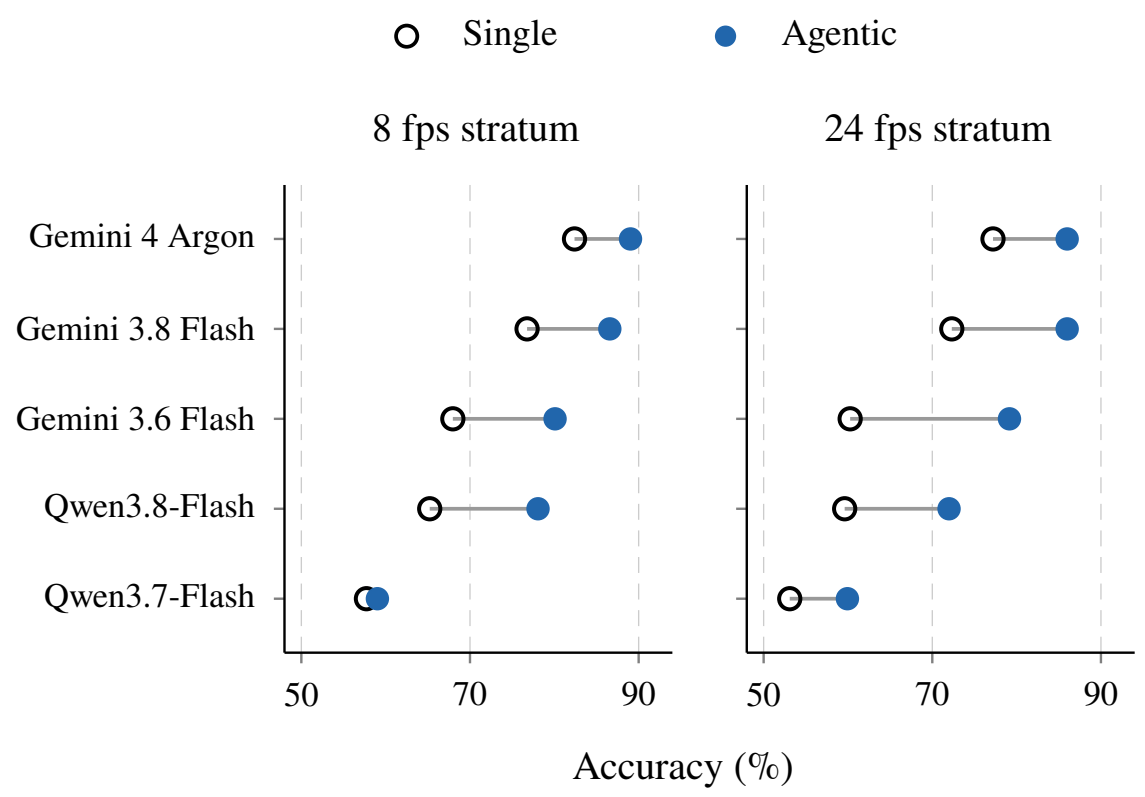
\begin{figure}[t]
\centering
\pgfplotsset{aceaxis/.style={scale only axis,axis lines*=left,axis line style={line width=.45pt},tick style={line width=.45pt},tick align=outside,major tick length=2pt,font=\fontsize{7}{8}\selectfont,label style={font=\fontsize{7.5}{9}\selectfont},tick label style={font=\fontsize{6.5}{8}\selectfont},grid style={gray!40,densely dashed,line width=.25pt},clip=false},acebar/.style={draw=black,line width=.3pt}}
\begin{tikzpicture}[x=1pt,y=1pt]
\path[use as bounding box] (0,0) rectangle (234,151);
\begin{axis}[aceaxis,at={(68pt,28pt)},anchor=south west,width=73pt,height=88pt,xmin=48,xmax=94,ymin=-.6,ymax=4.6,y dir=reverse,xtick={50,70,90},ytick={0,...,4},xmajorgrids,yticklabels={Gemini 4 Argon,Gemini 3.8 Flash,Gemini 3.6 Flash,Qwen3.8-Flash,Qwen3.7-Flash}]
\addplot[gray!80,line width=.7pt] coordinates {(82.395382,0) (89.033189,0)};
\addplot[only marks,mark=o,mark size=2pt,mark options={draw=black,fill=white,line width=.75pt}] coordinates {(82.395382,0)};
\addplot[only marks,mark=*,mark size=2pt,mark options={draw=aceblue,fill=aceblue}] coordinates {(89.033189,0)};
\addplot[gray!80,line width=.7pt] coordinates {(76.767677,1) (86.580087,1)};
\addplot[only marks,mark=o,mark size=2pt,mark options={draw=black,fill=white,line width=.75pt}] coordinates {(76.767677,1)};
\addplot[only marks,mark=*,mark size=2pt,mark options={draw=aceblue,fill=aceblue}] coordinates {(86.580087,1)};
\addplot[gray!80,line width=.7pt] coordinates {(67.965368,2) (80.08658,2)};
\addplot[only marks,mark=o,mark size=2pt,mark options={draw=black,fill=white,line width=.75pt}] coordinates {(67.965368,2)};
\addplot[only marks,mark=*,mark size=2pt,mark options={draw=aceblue,fill=aceblue}] coordinates {(80.08658,2)};
\addplot[gray!80,line width=.7pt] coordinates {(65.223665,3) (78.066378,3)};
\addplot[only marks,mark=o,mark size=2pt,mark options={draw=black,fill=white,line width=.75pt}] coordinates {(65.223665,3)};
\addplot[only marks,mark=*,mark size=2pt,mark options={draw=aceblue,fill=aceblue}] coordinates {(78.066378,3)};
\addplot[gray!80,line width=.7pt] coordinates {(57.720058,4) (59.018759,4)};
\addplot[only marks,mark=o,mark size=2pt,mark options={draw=black,fill=white,line width=.75pt}] coordinates {(57.720058,4)};
\addplot[only marks,mark=*,mark size=2pt,mark options={draw=aceblue,fill=aceblue}] coordinates {(59.018759,4)};
\end{axis}
\node[anchor=center,font=\fontsize{7.5}{8.5}\selectfont] at (104,126) {8 fps stratum};
\begin{axis}[aceaxis,at={(155pt,28pt)},anchor=south west,width=73pt,height=88pt,xmin=48,xmax=94,ymin=-.6,ymax=4.6,y dir=reverse,xtick={50,70,90},ytick={0,...,4},xmajorgrids,yticklabels={,,,,}]
\addplot[gray!80,line width=.7pt] coordinates {(77.198697,0) (85.993485,0)};
\addplot[only marks,mark=o,mark size=2pt,mark options={draw=black,fill=white,line width=.75pt}] coordinates {(77.198697,0)};
\addplot[only marks,mark=*,mark size=2pt,mark options={draw=aceblue,fill=aceblue}] coordinates {(85.993485,0)};
\addplot[gray!80,line width=.7pt] coordinates {(72.312704,1) (85.993485,1)};
\addplot[only marks,mark=o,mark size=2pt,mark options={draw=black,fill=white,line width=.75pt}] coordinates {(72.312704,1)};
\addplot[only marks,mark=*,mark size=2pt,mark options={draw=aceblue,fill=aceblue}] coordinates {(85.993485,1)};
\addplot[gray!80,line width=.7pt] coordinates {(60.260586,2) (79.153094,2)};
\addplot[only marks,mark=o,mark size=2pt,mark options={draw=black,fill=white,line width=.75pt}] coordinates {(60.260586,2)};
\addplot[only marks,mark=*,mark size=2pt,mark options={draw=aceblue,fill=aceblue}] coordinates {(79.153094,2)};
\addplot[gray!80,line width=.7pt] coordinates {(59.609121,3) (71.986971,3)};
\addplot[only marks,mark=o,mark size=2pt,mark options={draw=black,fill=white,line width=.75pt}] coordinates {(59.609121,3)};
\addplot[only marks,mark=*,mark size=2pt,mark options={draw=aceblue,fill=aceblue}] coordinates {(71.986971,3)};
\addplot[gray!80,line width=.7pt] coordinates {(53.094463,4) (59.934853,4)};
\addplot[only marks,mark=o,mark size=2pt,mark options={draw=black,fill=white,line width=.75pt}] coordinates {(53.094463,4)};
\addplot[only marks,mark=*,mark size=2pt,mark options={draw=aceblue,fill=aceblue}] coordinates {(59.934853,4)};
\end{axis}
\node[anchor=center,font=\fontsize{7.5}{8.5}\selectfont] at (191,126) {24 fps stratum};
\draw[black,line width=.7pt,fill=white] (91,144) circle (2pt);\node[anchor=west,font=\fontsize{7}{8}\selectfont] at (98,144) {Single};
\fill[aceblue] (151,144) circle (2pt);\node[anchor=west,font=\fontsize{7}{8}\selectfont] at (158,144) {Agentic};
\node[anchor=center,font=\fontsize{7.5}{8.5}\selectfont] at (143,6) {Accuracy (\%)};
\end{tikzpicture}
\caption{Accuracy on the temporal-density strata. Each segment connects single-pass accuracy (open circle) to autonomous tool-use accuracy (filled circle).}
\label{fig:fps}
\end{figure}

\begin{figure}[t]
\centering
\includegraphics[width=\columnwidth]{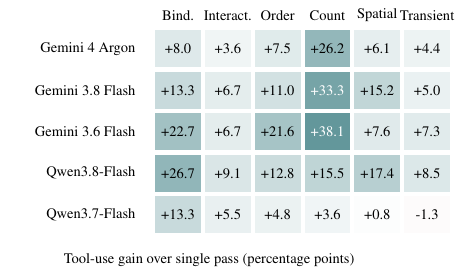}
\caption{Autonomous tool-use gains by action task, in percentage points. Each cell uses all questions of that type; sample sizes appear in \cref{tab:types}. Color is centered at zero and numerical values retain the sign.}
\label{fig:task-gains}
\end{figure}

\begin{figure*}[t]
\centering
\includegraphics[width=\textwidth]{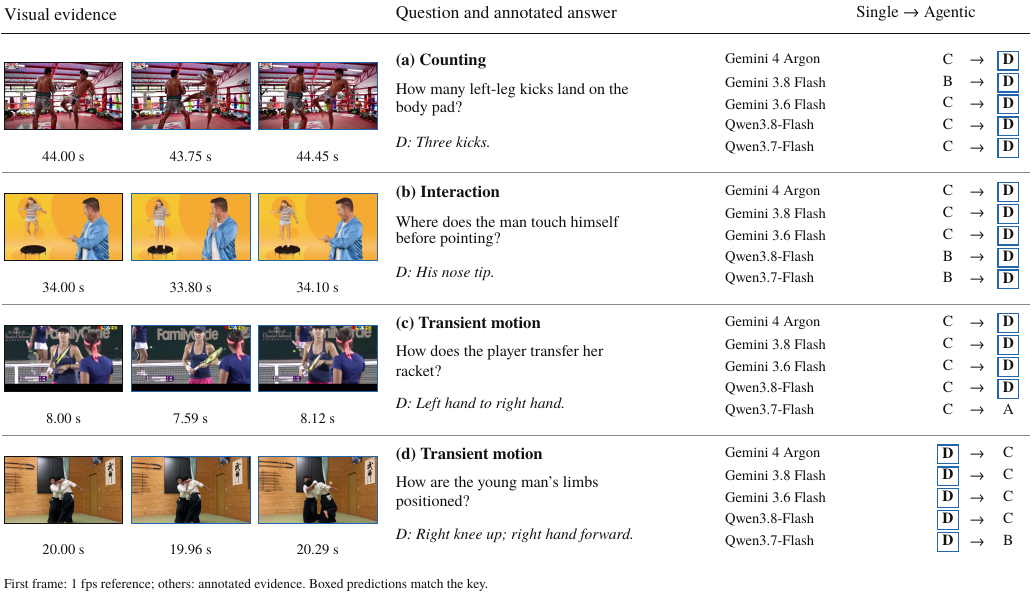}
\caption{Selected successes and regressions. Four selected questions with real source frames and predictions from all five models. The first two are fixed by every model, the third by four models, and the fourth is broken by all five. English text paraphrases the Chinese questions and selected options. Frames are taken from the annotated evidence. Boxed predictions match the annotated key.}
\label{fig:qual}
\end{figure*}

All five models improve when given access to further observations, by different amounts under the same tool interface. Gemini 3.6 Flash gains 14.2 points, from 65.6\% to 79.8\%, and Qwen3.8-Flash gains 12.7, whereas Qwen3.7-Flash gains 3.0. Gemini 4 Argon, which starts from the highest single-pass accuracy of 80.8\%, gains 7.3.

Models also use the round budget differently. Qwen3.7-Flash stops after one round in 55.7\% of groups, whereas Qwen3.8-Flash uses all three rounds in 66.3\% of groups and Gemini 3.8 Flash in 49.0\%. \Cref{tab:cost} reports full round counts, costs, and first-round agentic accuracy.

\subsection{Interaction on fine-grained action tasks}
\label{sec:breakdown}

For every model, the 307 questions annotated at 24\,fps have lower single-pass accuracy than the 693 annotated at 8\,fps (\cref{fig:fps}). Four models gain more on the 24\,fps stratum, and Qwen3.8-Flash gains about the same on both. Gemini 3.6 Flash improves by 18.9 points on the 24\,fps questions and 12.1 points on the 8\,fps questions. Questions that call for denser sampling benefit more from further observation.

Gains vary across tasks (\cref{fig:task-gains}; accuracies in \cref{tab:type}). Counting is the hardest type under single-pass observation, at 36.9--45.2\%, and yields the largest gain for the three Gemini models, since it requires resolving repeated events across time; tool use raises Gemini 3.6 Flash from 36.9\% to 75.0\% and Gemini 3.8 Flash from 42.9\% to 76.2\%. Qwen3.7-Flash gains only 3.6 points on counting and falls below its single-pass score on transient details, while Qwen3.8-Flash gains most on binding (26.7 points). Counting therefore separates models most clearly by how well they use additional observations.

\paragraph{An interaction example.}
\label{sec:qualitative}
In the Messi example of \cref{fig:teaser}, the model requests a dense clip and then selected frames before confirming its answer.

\paragraph{What additional observations resolve.}
Figure~\ref{fig:qual} links benchmark predictions to the visual distinctions the questions ask for. After tool use, all five models answer the kick-count and brief nose-touch examples correctly, and four recover the direction of a racket transfer. In the grappling-pose example, the single-pass answers match the key and the agentic answers do not.

\begin{figure*}[t]
\centering
\includegraphics[width=\textwidth]{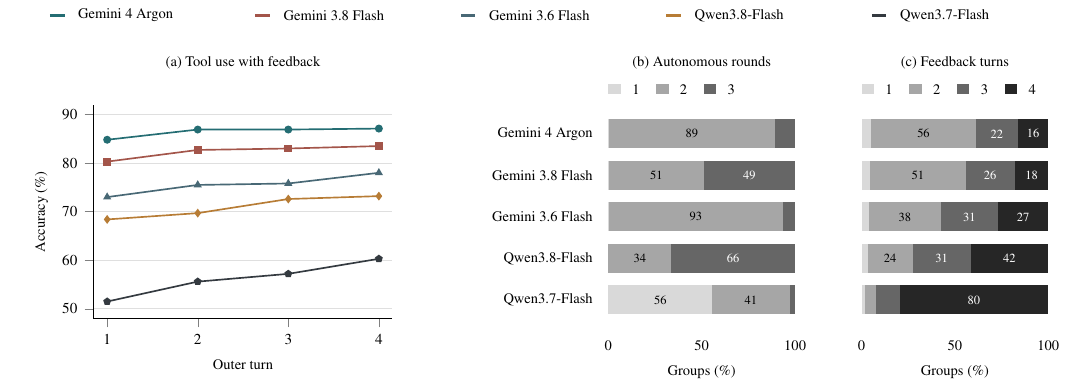}
\caption{Performance and use of interaction opportunities. (a) Accuracy through the combined tool-and-feedback episode, always over 1,000 questions; early-ending episodes carry their final answer forward. (b) Inner-round usage in autonomous tool sessions. (c) Outer-turn usage with feedback. Each distribution covers 300 groups; segment labels are rounded percentages, and narrow segments are unlabeled. Inner rounds and outer turns are distinct units.}
\label{fig:interaction-rounds}
\end{figure*}

\subsection{Tool use with simulated user feedback}
\label{sec:guided}

In the combined setting, the model inspects the video again each time new feedback arrives, with the tool backend of the autonomous experiments and the protocol of \cref{sec:guided-protocol}.

\begin{table}[t]
\centering\small
\caption{\textbf{Tool use with user feedback.} First and final refer to outer turns, each containing an inner tool session. All models answer the full 1,000 questions.}
\label{tab:combined-feedback}
\setlength{\tabcolsep}{4pt}
\begin{tabular}{@{}lrrr@{}}
\toprule
Model & First (\%) & Final (\%) & Gain \\
\midrule
Gemini 4 Argon & 84.8 & 87.1 & +2.3 \\
Gemini 3.8 Flash & 80.3 & 83.5 & +3.2 \\
Gemini 3.6 Flash & 73.0 & 78.0 & +5.0 \\
Qwen3.8-Flash & 68.4 & 73.2 & +4.8 \\
Qwen3.7-Flash & 51.5 & 60.3 & +8.8 \\
\bottomrule
\end{tabular}
\end{table}

Every model improves over its first outer turn, by 2.3 (Gemini 4 Argon) to 8.8 points (Qwen3.7-Flash), so feedback adds to the first tool session.

\paragraph{How improvement develops through interaction.}
Figure~\ref{fig:interaction-rounds}(a) follows the same 1,000 questions through each run. Gemini 4 Argon improves from 84.8\% to 86.9\% by turn two and finishes at 87.1\%. Qwen3.8-Flash keeps improving through turn four, from 68.4\% through 69.7\% and 72.6\% to 73.2\%, and Qwen3.7-Flash from 51.5\% through 55.6\% and 57.2\% to 60.3\%.

\paragraph{Interaction depth.}
In Figure~\ref{fig:interaction-rounds}(b,c), Qwen3.7-Flash ends 167 of 300 autonomous groups after one inner round, yet 239 of its feedback episodes reach four outer turns, against 49 for Gemini 4 Argon.

\subsection{Benchmark design and data quality}
\label{sec:design-quality}

To show how construction choices shape the evidence a benchmark asks for, we compare VideoACE with \emph{vanilla}, an independently constructed set of 1,000 questions on the same 100 source videos that covers broader actor binding, localization, and outcomes (\cref{tab:revision-quality}; Appendix~\ref{sec:supp-pipeline}). The median evidence duration is 0.90\,s for VideoACE against 3.00\,s for vanilla, and VideoACE contains 317 transient-detail and 307 24\,fps-annotated questions against 50 and 34. VideoACE thus concentrates evaluation on brief intermediate actions that call for targeted inspection, while option refinement brings both sets to a similar option-length spread of 0.038 and 0.037. In VideoACE, refinement also lowers the strictly-longest-key rate from 42.7\% to 8.5\%.

\begin{table}[t]
\centering\small
\caption{Task-design and option-quality statistics for two independently constructed question sets. Each has 1,000 items; source-video overlap is 100/100. Arrows compare options before and after refinement within each set.}
\label{tab:revision-quality}
\begin{tabular}{@{}lrr@{}}
\toprule
Measure & vanilla & VideoACE \\
\midrule
Median evidence duration (s) & 3.00 & 0.90 \\
24 fps annotations & 34 & 307 \\
Transient-detail questions & 50 & 317 \\
Strictly longest key (\%) & $32.6\to6.3$ & $42.7\to8.5$ \\
Mean relative length spread & $.203\to.037$ & $.205\to.038$ \\
Final keys per letter & 250 & 250 \\
\bottomrule
\end{tabular}
\end{table}

\section{Conclusion}
\label{sec:conclusion}

Agentic video understanding requires models to gather evidence by interacting with the video. \bench evaluates this capability on fast, fine-grained actions with 1,000 questions across six categories, each grounded in an annotated evidence interval. In its environment, video tools let models adjust temporal coverage, sampling density, and spatial detail, and a User Simulator prompts further inspection with leakage-screened hints.

Across single-pass observation, autonomous tool use, and tool use with user feedback, the five evaluated models show distinct patterns of interaction. Under the same interface, Gemini 3.6 Flash gains 14.2 points from tool use, while Qwen3.7-Flash gains 3.0. The two also differ in how far they pursue interaction: Qwen3.7-Flash stops after the first round in 55.7\% of groups, whereas Qwen3.8-Flash uses all three rounds in 66.3\%. Gains concentrate on questions that require detailed observation; Gemini 3.6 Flash raises its counting accuracy from 36.9\% to 75.0\%. Simulated feedback improves every model beyond its first tool session: Qwen3.7-Flash gains 8.8 points, yet its final accuracy of 60.3\% remains the lowest of the five. These results show that the use of interaction is a capability in its own right, which \bench measures.

\section*{LLM Usage Statement}
LLMs were used in this work only in the following, limited ways. In writing, they served in an assistive capacity: authors drafted the text and used LLMs to edit and refine it. In code and implementation, they likewise assisted with editing and refining author-written code. In data preparation, LLMs were used to generate part of the data (synthetic data, dataset creation, or annotation), and the authors reviewed, edited, and approved all generated output. LLMs were not used for related work and citations, for the design, execution, or analysis of experiments, for figures and visual material, or for research ideas, hypotheses, claims, and conclusions. Neither the full manuscript nor major portions of it were shared with any LLM provider to obtain reviewer-style feedback.

\FloatBarrier
\bibliographystyle{plainnat}
\bibliography{main}
\clearpage
\appendix
\setcounter{table}{0}
\setcounter{figure}{0}
\renewcommand{\thetable}{A\arabic{table}}
\renewcommand{\thefigure}{A\arabic{figure}}
\renewcommand{\theHtable}{appendix.\arabic{table}}
\renewcommand{\theHfigure}{appendix.\arabic{figure}}
\section{Collection, task design, and quality control}
\label{sec:supp-pipeline}

\paragraph{Candidate retrieval and provenance.}
The source collection uses category-specific searches for fights, sports, dance, dramatic action, single-person motion, and animation/game footage. The search script deduplicates video identifiers, constrains source duration to 30--300\,s, and excludes title patterns such as tutorials, interviews, trailers, and compilations. Supplemental retrieval artifacts also exist. The stored 1,015 candidate rows and 1,214 selected/download records come from different collection stages and batches. The final library contains 800 distinct source identifiers. Source categories come from retrieval metadata.

\paragraph{Selecting observable action windows.}
The window selector considers successfully probed videos of at least 720p. At 2\,fps, a person detector counts boxes with confidence at least 0.4 and area at least 0.4\% of the frame. Motion is the mean absolute difference between consecutive 160-by-90 grayscale samples, in the 0--255 intensity scale. A candidate window lasts at most 59.5\,s and receives score
\begin{equation}
0.6p+0.4\min(m/15,1),
\end{equation}
where $p$ is the fraction of samples meeting the category's person requirement and $m$ is mean motion. Multi-person categories require at least two visible people in 50\% of samples; single-person categories require one in 70\%. Animation/game footage bypasses person-count rejection because the detector is unreliable there. All categories require mean motion at least 1.5.

Window boundaries are adjusted toward nearby detected cuts within 5\,s. Extraction begins at a usable keyframe and uses stream copying; the exporter checks the resulting duration, shortening and retrying if necessary. Category quotas and a cap of six selected clips per channel constrain concentration. The stored rejection file contains 177 entries: 168 for insufficient multi-person occupancy, eight for insufficient single-person occupancy, and one for low motion.

\paragraph{Reproducible benchmark sampling.}
The sampling manifest records 800 source rows, a sample size of 100, seed 20261102, and hashes of both the source list and sample. Largest-remainder allocation preserves category proportions; seeded sampling is performed within each category. Stable identifiers y001--y100 retain the original video identifier, metadata row, temporal offset, and clip hash. The fineaction run reuses the source sample and prepares silent 24\,fps video. The two question sets share source identifiers but differ in their generation tasks and audio conditions.

\paragraph{Two independently constructed question sets.}
vanilla contains 307 binding, 133 interaction, 115 order, 105 counting, 118 spatial, 100 temporal, 50 transient, and 72 outcome questions. Its frame-rate annotations span 1/2/4/8/24\,fps, with 539 items below 8\,fps. Fineaction instead concentrates on six action-detail types and places all 1,000 annotations at 8 or 24\,fps. Its generation instructions require plausible options sharing coarse entry and exit states, differing in the intermediate action. The generator must check its own caption against the video and record corrections; it also checks a weaker caption for candidate mistakes. Actor identifiers from the two captions are not assumed to align: visible appearance is used to identify participants. These instructions aim to make the alternatives visually discriminative.

\paragraph{What option-quality improvements establish.}
We recompute Table~\ref{tab:revision-quality} from the saved original options/keys and their rewritten counterparts. A key is ``strictly longest'' only if its Unicode-character length exceeds every distractor; ties do not count. Relative spread is the range of the four option lengths divided by their mean, averaged over questions. Only 258 vanilla and 317 fineaction items have a unique longest option after rewriting. Under a uniformly assigned key, their expected strictly-longest-key rates are therefore 6.45\% and 7.93\%, respectively, close to the observed 6.3\% and 8.5\%. Balanced answer positions remove marginal letter imbalance.

\section{Complete autonomous results}

The following tables complement the main comparison. All entries use the same 1,000 questions and 300 evaluation groups. Table~\ref{tab:type} reports the six action skills; Table~\ref{tab:fps} reports the annotated temporal-density strata. Each pair gives single-pass and final agentic accuracy.

\begin{table*}[t]
\centering
\small
\setlength{\tabcolsep}{3pt}
\renewcommand{\arraystretch}{1.15}
\begin{tabular}{lrccccc}
\toprule
Type & $n$ & Gemini 4 Argon & Gemini 3.8 Flash & Gemini 3.6 Flash & Qwen3.8-Flash & Qwen3.7-Flash \\
\midrule
binding & 75 & 80.0 $\to$ 88.0 & 78.7 $\to$ 92.0 & 61.3 $\to$ 84.0 & 52.0 $\to$ 78.7 & 44.0 $\to$ 57.3 \\
interaction & 165 & 86.7 $\to$ 90.3 & 81.2 $\to$ 87.9 & 71.5 $\to$ 78.2 & 66.7 $\to$ 75.8 & 53.9 $\to$ 59.4 \\
order & 227 & 84.1 $\to$ 91.6 & 80.2 $\to$ 91.2 & 65.2 $\to$ 86.8 & 65.2 $\to$ 78.0 & 63.9 $\to$ 68.7 \\
counting & 84 & 42.9 $\to$ 69.0 & 42.9 $\to$ 76.2 & 36.9 $\to$ 75.0 & 45.2 $\to$ 60.7 & 42.9 $\to$ 46.4 \\
spatial & 132 & 84.1 $\to$ 90.2 & 68.9 $\to$ 84.1 & 65.9 $\to$ 73.5 & 60.6 $\to$ 78.0 & 57.6 $\to$ 58.3 \\
transient & 317 & 84.2 $\to$ 88.6 & 79.5 $\to$ 84.5 & 71.3 $\to$ 78.5 & 69.4 $\to$ 77.9 & 58.0 $\to$ 56.8 \\
\bottomrule
\end{tabular}
\caption{Accuracy (\%), \single $\to$ \relook, by question type on fineaction.}
\label{tab:type}
\end{table*}

\begin{table*}[t]
\centering
\small
\setlength{\tabcolsep}{3pt}
\renewcommand{\arraystretch}{1.15}
\begin{tabular}{lrccccc}
\toprule
$f_{\min}$ & $n$ & Gemini 4 Argon & Gemini 3.8 Flash & Gemini 3.6 Flash & Qwen3.8-Flash & Qwen3.7-Flash \\
\midrule
8 & 693 & 82.4 $\to$ 89.0 & 76.8 $\to$ 86.6 & 68.0 $\to$ 80.1 & 65.2 $\to$ 78.1 & 57.7 $\to$ 59.0 \\
24 & 307 & 77.2 $\to$ 86.0 & 72.3 $\to$ 86.0 & 60.3 $\to$ 79.2 & 59.6 $\to$ 72.0 & 53.1 $\to$ 59.9 \\
\bottomrule
\end{tabular}
\caption{Accuracy (\%), \single $\to$ \relook, by model-annotated minimum frame rate. These labels are not exposed to the evaluated model.}
\label{tab:fps}
\end{table*}

\section{Execution and accounting}

\paragraph{Inputs and outputs.}
Both conditions use the same question stems and options, including temporal references in the stems. Questions are grouped 4/3/3 per clip. Options are serialized on separate lettered lines, and the scorer parses four-choice answers. Each model has a valid final letter for all 1,000 items. First-round blanks in Table~\ref{tab:cost} denote deferral within the agentic run, not missing final predictions.

\paragraph{Tool limits.}
The agent begins with the complete clip at 1\,fps and has at most three answer rounds. Up to six observation requests are permitted in each non-final round. Dense clips are requested at up to 24\,fps and adjusted to a soft 120-frame budget; selected-frame requests return at most eight images. Crops, occurrence montages, 2\,fps scans, and refreshed overviews share the same executor. Conversation history is retained when new observations are appended.

\paragraph{Backend adaptations.}
Gemini receives the requested video frame rate through media metadata. Qwen requests above its API's 10\,fps limit are represented as slowed clips preserving the selected source frames. Consequently, source evidence is shared while its encoding depends on the backend. Gemini uses high reasoning effort. Qwen uses the gateway's default reasoning behavior, temperature 0.1, and a maximum of 8,192 output tokens.

\paragraph{Cost.}
Table~\ref{tab:cost} uses logged token usage and configured API price estimates for completed records, including the evaluated model's planning and answering calls. The five completed agentic evaluations total \$130.61. The table reports cost per complete 1,000-question evaluation.

\section{User Simulator details and supplementary chat results}
\label{sec:supp-guided}

\subsection{Role of the user simulator}
The User Simulator supplies task-relevant feedback within the interactive evaluation environment. It uses prepared evidence annotations to provide temporal and visual guidance for subsequent inspection. Its online hint policy is deterministic given the task artifact and the model's reported intervals; no additional language-model call selects a hint during an episode.

\subsection{Hint construction and release}
Let $E_i=[s_i,e_i]$ be the annotated evidence interval, with length $\ell_i$. The model reports up to three supporting windows $W$. A question is declared located if at least one window satisfies
\begin{equation}
\frac{|W\cap E_i|}{\max(\ell_i,10^{-3})}\geq 0.5,
\qquad |W|\leq\max(10\,\mathrm{s},3\ell_i).
\end{equation}
The upper bound prevents arbitrarily long reported intervals from trivially covering the evidence.

If a question is not located and a position hint remains, the simulator releases the next position hint. H1 contains the evidence in a coarse window of nominal length $\max(12\,\mathrm{s},2\ell_i)$, with a question-ID-seeded random offset and outward integer-second rounding. It is omitted when the nominal span exceeds 60\% of the clip. H2 rounds the evidence interval outward to 0.5\,s boundaries. Both are suppressed if the answer options themselves contain temporal expressions.

Once a question is located, or its position hints are exhausted, the simulator releases H3 once if available. This cue directs attention to a participant, body part, or object. The evaluated model still decides how to use its visual input and what answer to give. No hint explicitly reports answer correctness. An episode ends when no question can receive a new hint, or after four outer turns. Each outer turn returns answers and evidence intervals for the whole question group.

\subsection{Leakage checks and task consistency}
The simulator task artifact derives from the same 1,000 questions as the main benchmark. Its options and answer keys match the primary question bank exactly. Stems are rewritten to remove explicit temporal references; the artifact's checker flags temporal expressions in two stems. The original stems remain unchanged in the main experiments.

H3 candidates are generated offline from the question, annotated answer, rationale, and distractor explanations. The builder compares three text-only answers without a hint with three using the candidate. A candidate is discarded if the correct count increases by at least two or if it contains a temporal expression. A second check considers the other questions in the group: it rejects cues that identify answer-specific wording or increase another question's text-only correct count by at least two. This matters because all questions share the same conversation. The stored task set retains H1/H2 for 999 questions and H3 for 682; 145 H3 candidates were rejected by the within-question check and another 173 by the cross-question check.

\subsection{Supplementary chat-only run}
We report an exploratory Qwen3.8-Flash run using a chat adapter that receives the video at 1\,fps along with the accumulated dialogue and new hints. It completes 297 groups and 990 questions; the failed groups are y011/g2, y025/g1, and y027/g2. All 1,000 stored first-turn input records were checked against the current derived task artifact and match its question text and options. The 990-question analysis uses only completed episodes.

\begin{table*}[t]
\begin{minipage}[t]{0.48\textwidth}
\centering\small
\caption{\textbf{Guided chat diagnostic, Qwen3.8-Flash.} A separate exploratory run on 990 questions; it is not a full-set autonomous score.}
\label{tab:guided-current}
\setlength{\tabcolsep}{5pt}
\renewcommand{\arraystretch}{1.12}
\begin{tabular}{@{}lr@{}}
\toprule
Measure & Result \\
\midrule
Completed groups / planned & 297 / 300 \\
Accuracy at turns 1 / 2 & 61.2 / 62.6\% \\
Accuracy at turns 3 / 4 & 63.9 / 64.5\% \\
First-to-final fixes / breaks & 110 / 77 \\
Self-localization before hints & 46.8\% \\
Final errors / ever located & 351 / 351 \\
Mean hints per question & 1.63 \\
Recorded cost (USD) & 9.62 \\
\bottomrule
\end{tabular}
\end{minipage}\hfill
\begin{minipage}[t]{0.48\textwidth}
\centering\small
\caption{\textbf{Answer changes after guidance.} Percentages are conditional on the prior answer's correctness. Counts $n$ refer to adjacent-turn transitions, not distinct questions.}
\label{tab:guided-transitions}
\setlength{\tabcolsep}{4pt}
\renewcommand{\arraystretch}{1.12}
\begin{tabular}{@{}lrrrr@{}}
\toprule
& \multicolumn{2}{c}{Previously wrong} & \multicolumn{2}{c}{Previously right} \\
\cmidrule(lr){2-3}\cmidrule(l){4-5}
Cue & $n$ & Fixed (\%) & $n$ & Broken (\%) \\
\midrule
None & 282 & 0.7 & 501 & 0.4 \\
Position & 385 & 17.7 & 554 & 8.3 \\
Visual focus & 243 & 32.5 & 431 & 15.8 \\
\bottomrule
\end{tabular}
\end{minipage}
\end{table*}

For aggregate turn curves, an episode that ends early carries its last answer forward. The first-to-final fixes and breaks in Table~\ref{tab:guided-current} compare answers within a guided episode; they differ from the independent-protocol comparison used in the main table. Self-localization is computed over the 989 completed questions with position hints. ``Ever located'' means that a reported interval satisfies the overlap rule in at least one turn, not necessarily in the final turn.

Cue groups differ in prior localization and episode history. A pointer can tell the chat adapter where the relevant action lies, but its 1\,fps input still lacks the intermediate frames.

\section{Additional answer-level analyses}
\label{sec:supp-changes}

The following analyses retain answer-level detail from the autonomous evaluation. They supplement the aggregate gains in the main text.

\subsection{Paired outcomes and inspection depth}
Let $S_i$ and $A_i$ indicate whether the single-pass and final agentic answers are correct. We report accuracy and two paired counts:
\begin{align}
 F &= \sum_{i=1}^{N}(1-S_i)A_i, &
 B &= \sum_{i=1}^{N}S_i(1-A_i), \label{eq:fixbreak}\\
 \Delta_{\mathrm{acc}} &= 100\,(F-B)/N. \label{eq:gain}
\end{align}
Here $F$ counts \emph{fixes}, $B$ counts \emph{breaks}, and $\Delta_{\mathrm{acc}}$ is the gain in percentage points. We additionally report first-round agentic accuracy, counting deferred answers as incorrect; the number of executed rounds per group; and recorded API cost.

\begin{figure*}[t]
\centering
\resizebox{\linewidth}{!}{\pgfplotsset{aceaxis/.style={scale only axis,axis lines*=left,axis line style={line width=.45pt},tick style={line width=.45pt},tick align=outside,major tick length=2pt,font=\fontsize{7}{8}\selectfont,label style={font=\fontsize{7.5}{9}\selectfont},tick label style={font=\fontsize{6.5}{8}\selectfont},grid style={gray!40,densely dashed,line width=.25pt},clip=false},acebar/.style={draw=black,line width=.3pt}}
\begin{tikzpicture}[x=1pt,y=1pt]
\path[use as bounding box] (0,0) rectangle (495,148);

\definecolor{acefix}{HTML}{246D73}
\definecolor{acebreak}{HTML}{A3554A}
\definecolor{aceroundone}{HTML}{DCE4E8}
\definecolor{aceroundtwo}{HTML}{8DA3AE}
\definecolor{aceroundthree}{HTML}{486875}
\begin{axis}[aceaxis,at={(70pt,27pt)},anchor=south west,width=188pt,height=88pt,xbar,bar width=7pt,xmin=-160,xmax=250,ymin=-.6,ymax=4.6,y dir=reverse,ytick={0,...,4},yticklabels={Gemini 4 Argon,Gemini 3.8 Flash,Gemini 3.6 Flash,Qwen3.8-Flash,Qwen3.7-Flash},xtick={-100,0,100,200},xticklabels={100,0,100,200},xlabel={Questions},xmajorgrids]
\addplot[draw=none,fill=acefix,bar shift=0pt] coordinates {(125,0)};
\addplot[draw=none,fill=acebreak,bar shift=0pt] coordinates {(-52,0)};
\node[anchor=west,font=\fontsize{7}{8}\selectfont,text=acefix] at (axis cs:129,0) {125};
\node[anchor=east,font=\fontsize{7}{8}\selectfont,text=acebreak] at (axis cs:-56,0) {52};
\addplot[draw=none,fill=acefix,bar shift=0pt] coordinates {(149,1)};
\addplot[draw=none,fill=acebreak,bar shift=0pt] coordinates {(-39,1)};
\node[anchor=west,font=\fontsize{7}{8}\selectfont,text=acefix] at (axis cs:153,1) {149};
\node[anchor=east,font=\fontsize{7}{8}\selectfont,text=acebreak] at (axis cs:-43,1) {39};
\addplot[draw=none,fill=acefix,bar shift=0pt] coordinates {(197,2)};
\addplot[draw=none,fill=acebreak,bar shift=0pt] coordinates {(-55,2)};
\node[anchor=west,font=\fontsize{7}{8}\selectfont,text=acefix] at (axis cs:201,2) {197};
\node[anchor=east,font=\fontsize{7}{8}\selectfont,text=acebreak] at (axis cs:-59,2) {55};
\addplot[draw=none,fill=acefix,bar shift=0pt] coordinates {(217,3)};
\addplot[draw=none,fill=acebreak,bar shift=0pt] coordinates {(-90,3)};
\node[anchor=west,font=\fontsize{7}{8}\selectfont,text=acefix] at (axis cs:221,3) {217};
\node[anchor=east,font=\fontsize{7}{8}\selectfont,text=acebreak] at (axis cs:-94,3) {90};
\addplot[draw=none,fill=acefix,bar shift=0pt] coordinates {(146,4)};
\addplot[draw=none,fill=acebreak,bar shift=0pt] coordinates {(-116,4)};
\node[anchor=west,font=\fontsize{7}{8}\selectfont,text=acefix] at (axis cs:150,4) {146};
\node[anchor=east,font=\fontsize{7}{8}\selectfont,text=acebreak] at (axis cs:-120,4) {116};
\draw[gray,line width=.5pt] (axis cs:0,-.6) -- (axis cs:0,4.6);
\end{axis}
\node[anchor=west,font=\fontsize{7}{8}\selectfont] at (82,125) {\textcolor{acebreak}{$\leftarrow$ Breaks}};
\node[anchor=west,font=\fontsize{7}{8}\selectfont] at (173,125) {\textcolor{acefix}{Fixes $\rightarrow$}};
\begin{axis}[aceaxis,at={(329pt,27pt)},anchor=south west,width=164pt,height=88pt,xbar stacked,bar width=7pt,xmin=0,xmax=100,ymin=-.6,ymax=4.6,y dir=reverse,ytick={0,...,4},yticklabels={,,,,},xtick={0,50,100},xlabel={Groups (\%)},xmajorgrids,legend style={draw=none,at={(.5,1.03)},anchor=south,legend columns=3,font=\fontsize{6.5}{8}\selectfont},legend cell align=left]
\addplot[draw=none,fill=aceroundone] coordinates {(0.33333333,0) (0.33333333,1) (0.33333333,2) (0,3) (55.666667,4)};
\addplot[draw=none,fill=aceroundtwo] coordinates {(88.666667,0) (50.666667,1) (93,2) (33.666667,3) (41.333333,4)};
\addplot[draw=none,fill=aceroundthree] coordinates {(11,0) (49,1) (6.6666667,2) (66.333333,3) (3,4)};
\legend{1 round,2 rounds,3 rounds}
\end{axis}
\node[text=black,font=\fontsize{6.5}{8}\selectfont] at (402.25333333333333,104.84615384615384) {89\%};
\node[text=black,font=\fontsize{6.5}{8}\selectfont] at (371.09333333333336,87.9230769230769) {51\%};
\node[text=white,font=\fontsize{6.5}{8}\selectfont] at (452.82,87.9230769230769) {49\%};
\node[text=black,font=\fontsize{6.5}{8}\selectfont] at (405.8066666666667,71.0) {93\%};
\node[text=black,font=\fontsize{6.5}{8}\selectfont] at (356.6066666666667,54.076923076923066) {34\%};
\node[text=white,font=\fontsize{6.5}{8}\selectfont] at (438.6066666666667,54.076923076923066) {66\%};
\node[text=black,font=\fontsize{6.5}{8}\selectfont] at (374.64666666666665,37.153846153846146) {56\%};
\node[text=black,font=\fontsize{6.5}{8}\selectfont] at (454.18666666666667,37.153846153846146) {41\%};
\node[anchor=west,font=\fontsize{8}{9}\selectfont] at (0,141) {(a) Corrections and regressions};
\node[anchor=west,font=\fontsize{8}{9}\selectfont] at (280,141) {(b) Executed answer rounds};
\end{tikzpicture}}
\caption{Paired answer changes and inspection depth. Left: fixes and breaks relative to the independent single-pass run; numbers are question counts. Right: executed answer rounds across 300 groups per model.}
\label{fig:behavior}
\end{figure*}
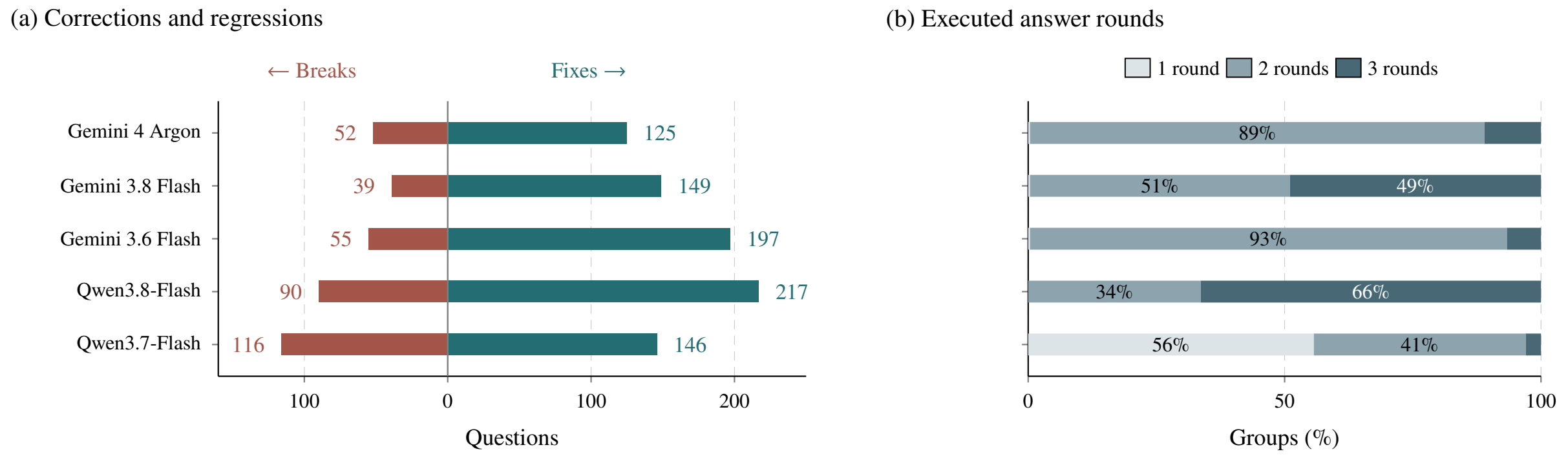
Qwen3.8-Flash corrects 217 baseline errors and changes 90 baseline-correct answers to incorrect ones. Qwen3.7-Flash corrects 146 and changes 116. Round usage in \cref{fig:behavior} provides an accompanying description of how much of the interaction budget each model uses.

\end{document}